\documentclass[conference]{IEEEtran}
\IEEEoverridecommandlockouts
\usepackage{cite}
\usepackage{amsmath,amssymb,amsfonts}
\usepackage{algorithmic}
\usepackage{graphicx}
\usepackage{textcomp}
\usepackage{xcolor}
\usepackage{hyperref}
\usepackage{subfig}
\usepackage{algorithm}
\usepackage{algorithmic}
\usepackage{pifont}
\newcommand{\cmark}{\ding{51}}%
\newcommand{\xmark}{\ding{55}}%
\makeatletter
\newcommand\fs@norules{\def\@fs@cfont{\bfseries}\let\@fs@capt\floatc@ruled
  \def\@fs@pre{}%
  \def\@fs@post{}%
  \def\@fs@mid{\kern3pt}%
  \let\@fs@iftopcapt\iftrue}
\makeatother
\floatstyle{norules}
\restylefloat{algorithm}

\def\BibTeX{{\rm B\kern-.05em{\sc i\kern-.025em b}\kern-.08em
    T\kern-.1667em\lower.7ex\hbox{E}\kern-.125emX}}
\begin{document}

\title{Learning and reusing primitive behaviours to improve Hindsight Experience Replay sample efficiency \\
\thanks{This publication has emanated from research supported by Science Foundation Ireland (SFI) under Grant Number SFI/12/RC/2289\_P2, co-funded by the European Regional Development Fund, by Science Foundation Ireland Future Research Leaders Award (17/FRL/4832), and by China Scholarship Council (CSC No.202006540003). For the purpose of Open Access, the author has applied a CC BY public copyright licence to any Author Accepted Manuscript version arising from this submission. We would like to express our deepest gratitude to Prof. Kevin McGuinness for his valuable contributions to this research. Unfortunately, he passed away before the completion of this work. He will be remembered for his exceptional insights and dedication.}
}

\author{\IEEEauthorblockN{Francisco Roldan Sanchez}
\IEEEauthorblockA{\textit{Dublin City University} \\
\textit{Insight SFI Centre for Data Analytics}\\
Dublin, Ireland \\
francisco.sanchez@insight-centre.org}
\and
\IEEEauthorblockN{Qiang Wang}
\IEEEauthorblockA{\textit{University College Dublin} \\
Dublin, Ireland \\
qiang.wang@ucdconnect.ie}
\and
\IEEEauthorblockN{David Cordova Bulens}
\IEEEauthorblockA{\textit{University College Dublin} \\
Dublin, Ireland \\
david.cordovabulens@ucd.ie}
\and
\IEEEauthorblockN{Kevin McGuinness}
\IEEEauthorblockA{\textit{Dublin City University} \\
\textit{Insight SFI Centre for Data Analytics}\\
Dublin, Ireland \\
kevin.mcguinness@insight-centre.org}
\and
\IEEEauthorblockN{ Stephen J. Redmond}
\IEEEauthorblockA{\textit{University College Dublin} \\
\textit{Insight SFI Centre for Data Analytics}\\
Dublin, Ireland \\
stephen.redmond@ucd.ie}
\and
\IEEEauthorblockN{ Noel E. O'Connor}
\IEEEauthorblockA{\textit{Dublin City University} \\
\textit{Insight SFI Centre for Data Analytics}\\
Dublin, Ireland \\
noel.oconnor@insight-centre.org}
}
\maketitle

\begin{abstract}
Hindsight Experience Replay (HER) is a technique used in reinforcement learning (RL) that has proven to be very efficient for training off-policy RL-based agents to solve goal-based robotic manipulation tasks using sparse rewards. Even though HER improves the sample efficiency of RL-based agents by learning from mistakes made in past experiences, it does not provide any guidance while exploring the environment. This leads to very large training times due to the volume of experience required to train an agent using this replay strategy. In this paper, we propose a method that uses primitive behaviours that have been previously learned to solve simple tasks in order to guide the agent toward more rewarding actions during exploration while learning other more complex tasks. This guidance, however, is not executed by a manually designed curriculum, but rather using a critic network to decide at each timestep whether or not to use the actions proposed by the previously-learned primitive policies. We evaluate our method by comparing its performance against HER and other more efficient variations of this algorithm in several block manipulation tasks. We demonstrate the agents can learn a successful policy faster when using our proposed method, both in terms of sample efficiency and computation time. Code is available at \url{https://github.com/franroldans/qmp-her}.
\end{abstract}


\section{Introduction}
Data-driven robotic manipulation has become very popular in recent years due to the success of reinforcement learning (RL) algorithms~\cite{mlrm}. By analysing large amounts of data, data-driven robotic manipulation methods allow the robot to carry out tasks with higher precision and efficiency~\cite{comparison}. Furthermore, because these techniques let robots learn from successes and mistakes, they can potentially be used to continuously improve the robot's abilities over time, leading to greater adaptability to different situations without the need of specialised programming~\cite{dqn}. However, to achieve such levels of adaptability, it is usually necessary to generate large quantities of good quality data, which leads to  prohibitively large training times for tasks and robots with complex observation, goal or/and action spaces. Therefore, training RL-based agents on such environments can be very challenging or even infeasible depending on the  computational resources available~\cite{her_results, tactile_her, hrl}. 

The most popular techniques used to solve data-driven robotic manipulation tasks are off-policy reinforcement learning methods, due to their sample efficiency and their ability to decouple exploration from exploitation \cite{ddpg, per, er}. Firstly, off-policy reinforcement learning algorithms are more sample efficient because they update their data after analysing a diverse set of trajectories, making them converge faster than when being immediately updated after each interaction with the environment. Secondly, exploration and exploitation are decoupled, meaning that one can use a different policy for exploration, termed a \textit{behaviour policy}, while the learning policy, termed a \textit{target policy}, prioritises exploiting what is learned from these well-chosen actions. One popular technique is Deep Deterministic Policy Gradients (DDPG) \cite{ddpg} in conjunction with Hindsight Experience Replay (HER) \cite{her}, which has proven to be very efficient for goal-based tasks with sparse rewards (i.e. binary rewards). 

This paper proposes a novel method that evaluates and selects behaviours generated by policies that are capable of performing simpler primitive skills. In this way we can accelerate the learning of new more skillful policies able to perform more difficult manipulation tasks by integrating this method of suggesting and selecting actions using primitive policies into the exploration-exploitation logic of the learning algorithm. We evaluate our method on four different robotic block manipulation tasks, comparing its success rate, its sample efficiency, and its computation time against an agent trained using the original HER algorithm and other more efficient variants of HER.

\section{Background}

\subsection{Q-learning}
\label{sec:q}

Q-learning \cite{q} is a popular off-policy RL algorithm used to train an agent to make optimal decisions in a Markov decision process environment. When training an agent using Q-learning, a $\mathbb{Q}$-table of state-action pairs must be preset by an initial exploration, along with the state-action pair Q-value. The Q-value is a metric that captures the quality of the transition, which is represented by a state $s$, an action $a$, the future state $s'$ reached by the taking the action, and a reward $r$. In other words, it defines how good or bad an action $a$ is that is taken for a particular state $s$, when following a policy~$\pi$. 

In particular, the Q-value for a state is the expected discounted cumulative reward given the state and the action. During training, as the environment is further explored, these Q-values must be updated based on the observed rewards using the Bellman equation \cite{bellman_eq}: 
 \begin{equation}
 \label{eq:update}
     Q^{\pi}_{new} = Q(s,a)+ \alpha [R(s,a) + \gamma \max_{a'} (Q(s', a'))  - Q(s, a)], 
 \end{equation}
 where $Q(s,a)$ denotes for the current Q-value, $\alpha$  is the learning rate, $R(s,a)$ is the reward for taking action $a$ when in state $s$, $\gamma$ is the discount factor,  determining the importance of future rewards, and $\max(Q(s', a'))$ is the maximum expected reward in the new state $s'$ given all possible actions that can be taken in the next state~$s'$.

\subsection{Deep Deterministic Policy Gradients and Hindsight Experience Replay}

DDPG \cite{ddpg} is an off-policy RL algorithm used for continuous control that is built using an actor-critic architecture, with an actor network $\pi(s)$ that predicts actions and a critic network $Q^{\pi}(s, a)$ that estimates the associated Q-value (a measure of quality) of pairs of states and actions. 

During training, the actor network updates its parameters to maximise the estimated Q-value, while the critic network is updated by minimising the mean squared temporal difference error \cite{td_error} between the estimated Q-value and the target Q-value, which is computed using equation~\ref{eq:update}. In order to perform these updates, data is sampled from a replay buffer that stores a diverse set of trajectories obtained from previous interactions with the environment (i.e. temporal sequences of environment transitions as defined in section~\ref{sec:q}).

HER \cite{her} is a replay strategy that accelerates the learning of off-policy RL algorithms by altering the stored data from past experiences. While in a standard RL framework an agent only can learn from trajectories receiving a positive reward, HER changes that by modifying the goals of failed stored experiences so that they match the achieved goal, thus obtaining a positive reward from what was a failed episode. In other words, it makes the agent learn from its own mistakes.

Despite the benefits of HER, this method requires a large quantity of simulated experience to find a good policy. The main reason for this requirement is the lack of guidance given to the agent during exploration \cite{hgg, ghgg}, particularly in the first interactions with the environment, as HER becomes most useful when the agent interacts with the object and changes its state, creating greater variety in the hindsight goals.

There are some algorithms that modify HER in order to tackle this problem. Hindsight Goal Generation (HGG) \cite{hgg} addresses the issue by generating a set of alternative goals based on the achieved states during the exploratory episodes. Alternatively, in \cite{ghgg}, a method named Graph-based Hindsight Goal Generation (G-HGG) is proposed to replace the Euclidean distance metric normally used to evaluate goal-based tasks by a graph-based distance metric, and in this way improve the selection of intermediate goals in environments with physical obstacles. However, both of these methods have very time-consuming policy updates, mostly because of the trajectory sampling required to use them, as for each generated goal, trajectories that match the generated goal states must be sampled from the replay buffer .

\subsection{Ensemble of policies}

An ensemble of policies is a collection of $K$ individual policies $ \mathbb{P} = \{ \widehat{\pi}_{1}, \widehat{\pi}_{2}, ...,  \widehat{\pi}_{K}\} $ that are used together to solve a task. Each policy in the ensemble $\mathbb{P}$ may have different strengths and weaknesses, and by combining their decisions, the ensemble can benefit from the range and variety of skills possessed by the policies that make up the ensemble, this improving the overall performance of the agent. 

There are different strategies to create such ensembles. For example, in \cite{maple}, the authors define a hierarchical policy, with a high-level task policy that decides which lower-level primitive policy of the ensemble to use, and a low-level parameter policy that determines how to instantiate the chosen primitive, in order to learn how to order the selection of these policies and their parameters over time.

SUNRISE \cite{sunr} is another ensemble mechanism that trains $N$ independent actor-critic based agents in a task and then selects the candidate action of the agent that maximises a \textit{weighted Bellman backup} equation, which is calculated using the mean and variance of the Q-values obtained by each candidate action using the critics of each of the agents of the ensemble. This idea of filtering actions based on the Q-value has also been used in different imitation learning frameworks and offline RL methods to filter noisy actions or unwanted behaviours~\cite{offline, bc}. 

A more recent method is the Q-switch Mixture of Policies (QMP) \cite{qsm}, which uses a Q-filter in a multi-task setting designed to simultaneously find policies that solve different tasks that share the observation space in order to identify shareable behaviours between policies. During exploration, each of the policies of the ensemble proposes a candidate action, but only the action that maximises the Q-value associated with the target policy for the task being solved is executed. 

In this paper, we use a QMP to improve the sample efficiency of HER, but instead of sharing behaviours between policies that are being simultaneously trained in the same environment, we make use of fixed primitive policies that do not necessarily share the observation or/and the goal spaces with the target task.

\section{Method}

The focus of this work is to improve the sample efficiency of DDPG agents trained with HER for robotic manipulation tasks by reusing simpler primitive skills that can be more easily learned. These primitive skills are policies obtained by training a DDPG agent using HER in simpler tasks. 

We define a set of $L$ objectives $\mathbb{O}= \{o_{1}, o_{2}, ..., o_{L}\}$ that correspond to potential state-goal combinations that $K$ primitive policies $\mathbb{P} = \{\widehat{\pi}_{1}(s, g), \widehat{\pi}_{2}(s, g), ..., \widehat{\pi}_{K}(s, g)\}$ could use. For each timestep during exploration, we use each primitive policy in $\mathbb{P}$ to create a candidate action for each of the objectives, and store them along with the candidate action $a_{t}^{\pi}$ that the target policy $\pi(s,g)$ proposes. After that, all candidate actions $\mathcal{A}_{t} = \{a_{t}^{\pi}, a_t^{1, 1}, a_t^{1,2}, ..., a_t^{K,L}\}$ for that timestep are evaluated by the critic of the target policy $Q^{\pi}(s, a)$, and only the action achieving the maximum Q-value for the current state~$s_t$ is executed and the transition stored, as done in \cite{qsm}:

\begin{equation}
\label{eq:maxq}
    a^* = arg \max Q^{\pi} (s_{t}, g, \mathcal{A}_{t}) 
\end{equation}

Before any update is performed, the Q-value that the critic predicts will be completely random, and as such during warm-up timesteps (i.e. timesteps of data collection before the actual learning process begins) the agent is forced to pick an action from the suggestions coming from the ensemble of primitive policies, along with the $\epsilon$-greedy algorithm (i.e., the agent occasionally explores the environment using random actions with a probability $\epsilon$). After the warm-up, the agent also considers the actions suggested by the target policy.

For all tasks, it is expected that the actions suggested by the primitive policies will mostly be used during the first iterations of the exploration, but as the target policy continues to update, the actions predicted by the target policy will have a higher Q-value. This means that by the end of the training, our method's behaviour will naturally evolve to be the same as original HER. 

We call this algorithm Q-switch Mixture of Primitives Hindsight Experience Replay (QMP-HER) - see Algorithm \ref{alg:qmp-her} for more details.

 \begin{algorithm}[t]
 \caption{Q-switch Mixture of Primitives Hindsight Experience Replay}
 \begin{algorithmic}[1]
  \label{alg:qmp-her}
 \renewcommand{\algorithmicrequire}{\textbf{Given:}}
 \REQUIRE  Off-policy RL agent $\mathbb{A} \sim  \pi (s, g) \; , Q^{\pi}(s, g, a)$. \\
  $K$ primitives: $\mathbb{P} = \{\widehat{\pi}_{1}(s, g), \widehat{\pi}_{2}(s, g), ... , \widehat{\pi}_{K}(s, g)\}$ \\
  Replay Buffer $\mathcal{R}$
  \STATE Initialise $\mathbb{A}$
  \STATE Load $\mathbb{P} $
    \STATE Define a set of $L$ objectives $\mathbb{O}= \{o_{1}, o_{2}, ... , o_{L}\}$ for $\mathbb{P}$

  \FOR{ epoch = 0 to $N$}
  \FOR {episode =1 to $M$}
  \STATE Sample a goal $g$ and an initial state $s_{0}$
  \FOR {t =0 to $T -1$}
   \FOR{$k=1$ to $K$ }\label{alg:start}
   \FOR{$l=1$ to $L$}
        \STATE  $s', g' \sim$ Adjust state and/or goal based on $o_{l}$
    \STATE Sample candidate action $a_{t}^{k, l}$ from $\widehat{\pi}_{k}(s', g')$
    \STATE $\mathcal{A}_{t} \sim$ Store candidate action $a_{t}^{k, l}$
  \ENDFOR
  \ENDFOR \label{alg:end}
  \STATE Sample candidate action $a_{t}^{\pi}$ from $\pi(s_{t}, g)$
  \STATE$\mathcal{A}_{t} \sim$ Store candidate action  $a_{t}^{\pi}$
\STATE Select best action $a^*$ using Eq. \ref{eq:maxq}, applying $\epsilon$-greedy
  \STATE Execute action $a^*$ and observe state $s_{t+1}$
    \STATE  $\mathcal{R} \sim$ Store transition $(a^*, s_{t}, s_{t+1}, g, r_{t})$ 
  \ENDFOR

  \ENDFOR
  \STATE Apply HER logic and update $\pi(s, g)$, $Q^{\pi}(s, g, a)$
  \ENDFOR
 \end{algorithmic}
 \end{algorithm}

\section{Experiments}
\label{sec:exp}
\subsection{Environment}

\begin{figure}[!t]

     \centering
  \subfloat[\label{1a}]{%
       \includegraphics[width=0.26\columnwidth]{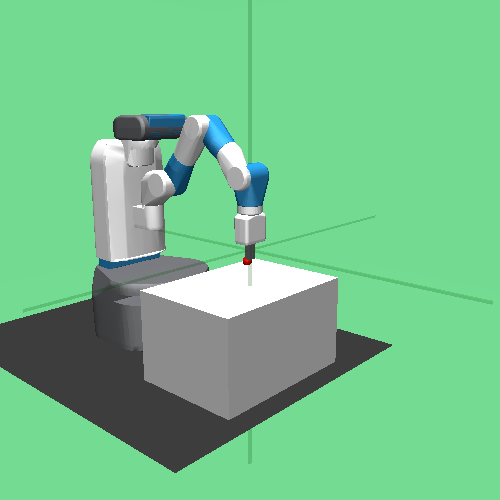}}
    \hfill
    \centering
  \subfloat[\label{1b}]{%
        \includegraphics[width=0.26\columnwidth]{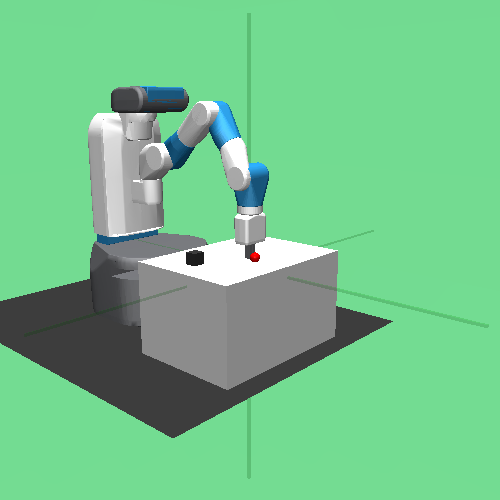}}
  \hfill
     \centering
  \subfloat[\label{1c}]{%
        \includegraphics[width=0.26\columnwidth]{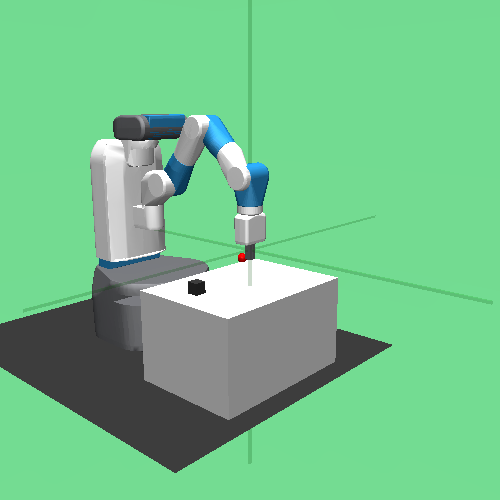}} \\
        \centering 
    \subfloat[\label{1d}]{%
        \includegraphics[width=0.26\columnwidth]{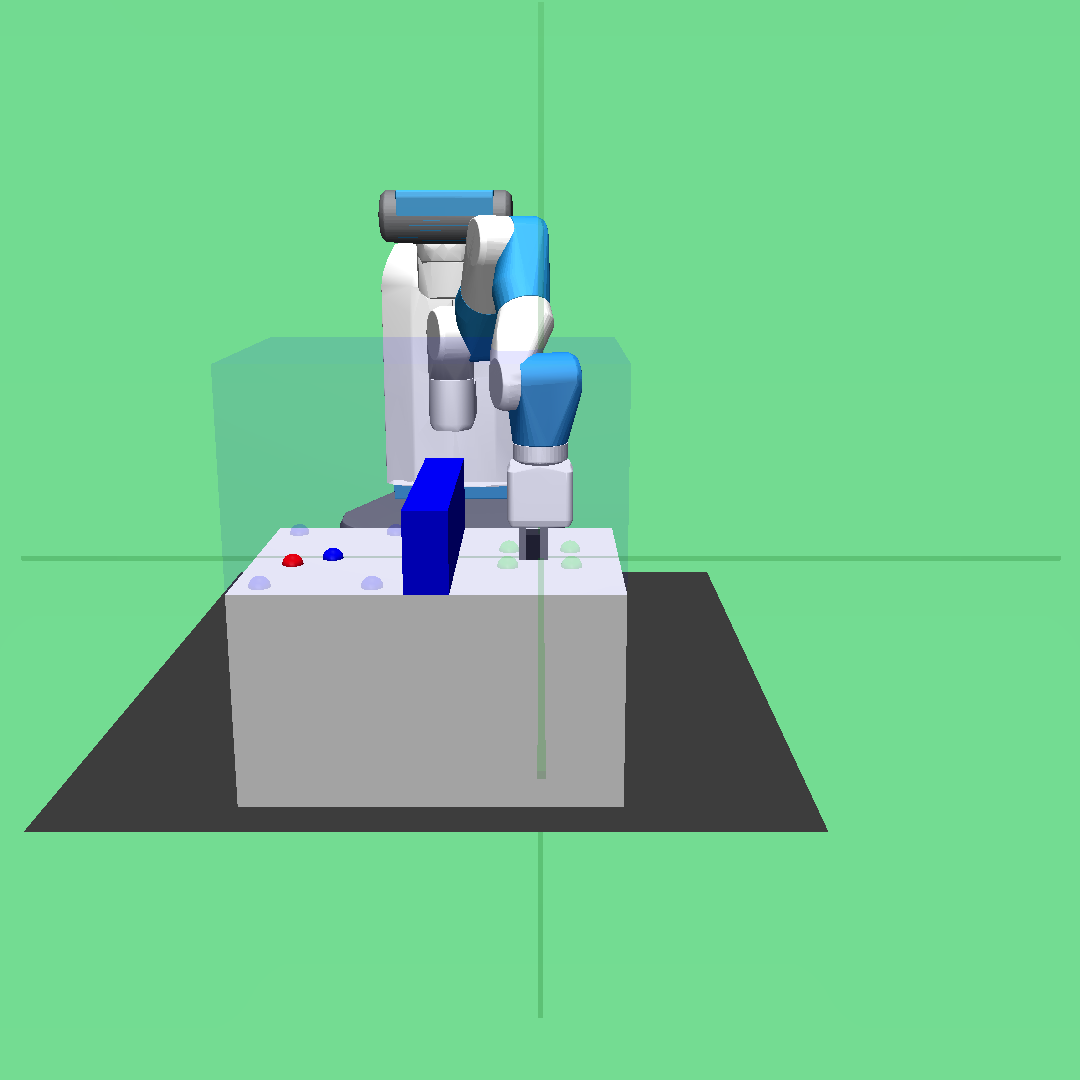}}
    \hspace{1 cm}
    \subfloat[\label{1e}]{%
        \includegraphics[width=0.26\columnwidth]{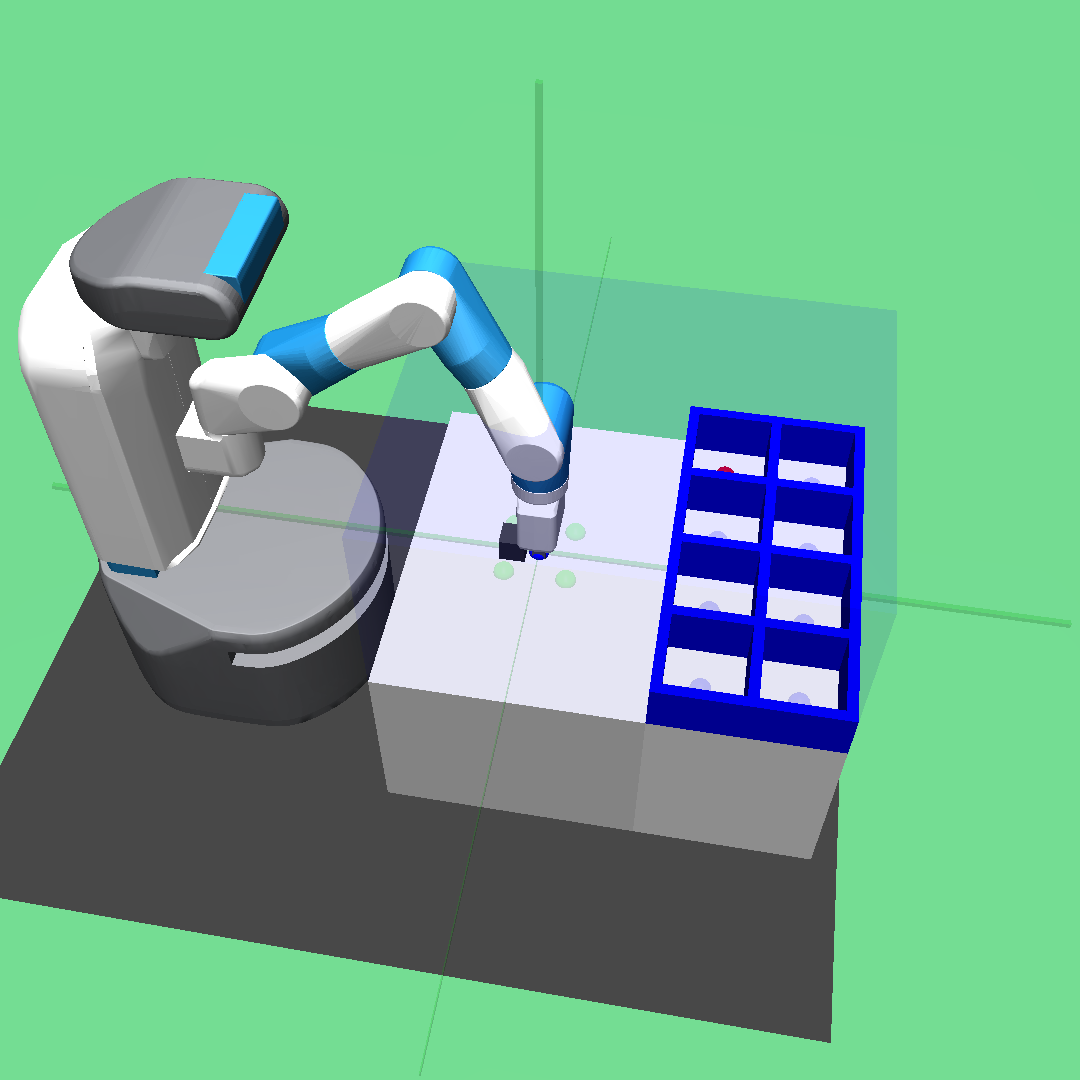}}
        \caption{Visual representation of (a)~\textit{FetchReach}, (b)~\textit{FetchPush}, (c)~\textit{FetchPickAndPlace}, (d)~\textit{FetchPickObstacle} and (e)~\textit{FetchPickAndThrow} tasks.} 
        \label{fig:envs}
\end{figure}

 We make use of the block manipulation tasks proposed by OpenAI for the Fetch robot arm using a parallel-jaw gripper~\cite{her_results}, as well as two more tasks proposed in~\cite{ghgg}. These are summarised as follows in order of complexity:

\begin{itemize}
    \item \textit{FetchReach:} the goal is to move the parallel-jaw gripper to a position in the 3D space. The gripper remains closed (Fig~\ref{1a}). 
    \item \textit{FetchPush:} the goal is to move a block to a point in the 3D space that is over the table. The gripper~remains closed (Fig~\ref{1b}).
    \item  \textit{FetchPickAndPlace:} the goal is to move a block to a point in the 3D space. Gripper controls are activated (Fig~\ref{1c}).
    \item \textit{FetchPickObstacle:} the goal is to move a block to a point in the 3D space with an obstacle between the initial and the target position. Gripper controls are activated (Fig~\ref{1d}).
    \item \textit{FetchPickAndThrow:} the goal is to lift a cube and throw it into one of eight boxes that are out of reach for the robotic arm. Gripper controls are activated (Fig~\ref{1e}).
\end{itemize}

\begin{table}[!t]
  \caption{Overview of the different environments observation and goal spaces}
  \label{scores1}
  \setlength{\tabcolsep}{2.5pt}
  \centering
  \begin{tabular}{c|cccc}
    Environment & Obs. Dimension & Object info. & Obstacles & Goal space \\
    \hline 
    \textit{FetchReach}  & 10  & \xmark & \xmark & Continuous  \\ 
    \textit{FetchPush}  & 25 & \cmark &  \xmark & Continuous \\
    \textit{FetchPickAndPlace}  & 25 & \cmark & \xmark & Continuous \\
    \textit{FetchPickObstacle}  & 25 & \cmark & \cmark & Continuous \\
    \textit{FetchPickAndThrow}  & 25 & \cmark & \cmark & Discrete \\

  \end{tabular}
\end{table}
To improve sample efficiency, we aim to make use of the behaviour learned by the agents trained on the simplest tasks. Therefore, the method we present in this paper cannot be used in the \textit{FetchReach} environment, as it is the simplest of all and it must be trained with HER. Once this policy is learned, there is then knowledge of the dynamics of the robot that has been acquired by this behaviour policy which can be used to more quickly learn how to perform other more complex tasks.

Because the observation space of the \textit{FetchReach} environment is smaller than the observation space of the other three environments, it is necessary to remove the object state information from the observations of these other environments in order to use the policy learned on the \textit{FetchReach} task.

\subsection{Experimental setup}

Our method requires a heuristic definition of the objectives for the primitive policies. These objectives are defined based on regions of interest in the 3D space relative to the object and the goal position, and might vary depending on the target task. For example, in order to learn a push or lifting policy, a robot can benefit from approaching the object from all possible directions. However, when learning a policy in tasks with physical obstacles, the robot benefits from manipulating the object in an elevated position, avoiding the obstacles.

To solve the \textit{FetchPush} task we define the objectives by setting a number of different goals for the \textit{FetchReach} primitives. For each observation of the \textit{FetchPush} task, we adjust it by removing the object information and request the \textit{FetchReach} primitive policy to move towards the location of the object and the goal position, but we also include objectives that point toward the object with slight deviations, so that the gripper touches the object from the left, right, front and behind. To solve the \textit{FetchPickAndPlace} task, we make use of the same primitives and objectives as the used to solve the textit{FetchPush} task. For this experiment, we also use the target policy during warm-up timesteps as it is necessary for the gripper state exploration because the policy trained on the \textit{FetchReach} task does not have the gripper controls activated.  

The \textit{FetchPickObstacle} is a task of sufficient complexity that HER cannot solve it. However, the task has many similarities with the \textit{FetchPickAndPlace} task and the policy learned in that task can be useful to accelerate the learning pace. We perform two different experiments: the first one being a manually designed curriculum which we implement by enforcing the behaviour policy to perform the first 20 timesteps of each episode using the policy learned on the \textit{FetchPickAndPlace} task with the goal to lift the object over the obstacle, the following 20 timesteps to move it towards the original goal, and for the remainder of the episode the agent executes the actions predicted by its actor network $\pi(s)$. We call this framework \textit{PickPlace\&HER}. In the second experiment, we use our method defining two objectives for the \textit{FetchPickAndPlace} primitive, one that aims to pick the block and move it above the obstacle and a second one aiming for the goal position.

The \textit{FetchPickAndThrow} task is much more challenging than the previous tasks, as the robot has to lift up the cube and throw it into a basket, meaning that the goal space is a discrete set of eight regions instead of the continuous goal space we had in previous experiments. Similar to the \textit{FetchPickObstacle} task, we believe that this task can be accelerated by starting with the object grasped by the gripper. Therefore, we run two different experiments, a first one running the same strategy as we defined for the \textit{PickPlace\&HER} exploration algorithm (i.e., grasping the object and moving it to a centred and elevated position, then letting the target policy explore). The second experiment involves our method using a combination of primitive behaviours: we use the policy learned on the \textit{FetchPickObstacle} task, setting as the objective the same goal that was sampled for that episode; and the policy learned on the \textit{FetchPickAndPlace} task, setting as objectives a centred and elevated position and the sampled goal for that episode. 

To calculate the success rate, we do not apply the QMP-HER logic (behaviour policy). Instead, we only test the actions predicted by the actor of the target policy $\pi(s)$ that is being updated to learn the execution of the task end-to-end. 

\section{Results}

\begin{figure*}[!t]

     \centering
  \subfloat[\label{3a}]{%
       \includegraphics[width=0.25\textwidth ]{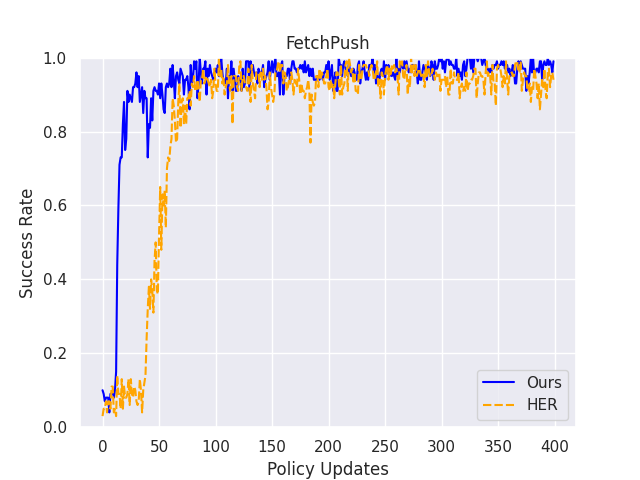}}
    \hfill
    \centering
  \subfloat[\label{3b}]{%
        \includegraphics[width=0.25\textwidth ]{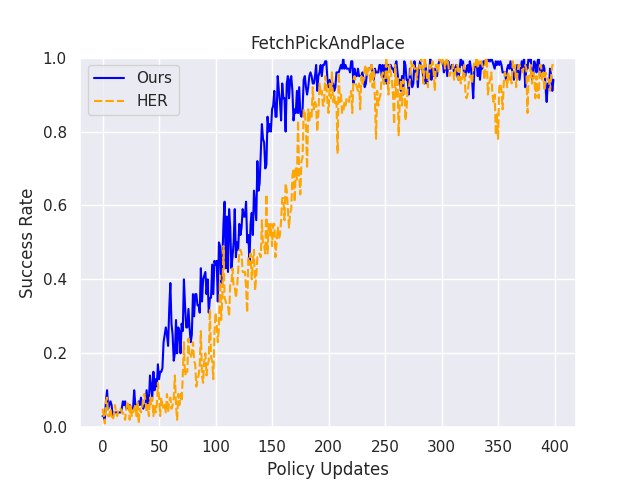}}
  \hfill
 \centering
    \subfloat[\label{3c}]{%
        \includegraphics[width=0.25\textwidth ]{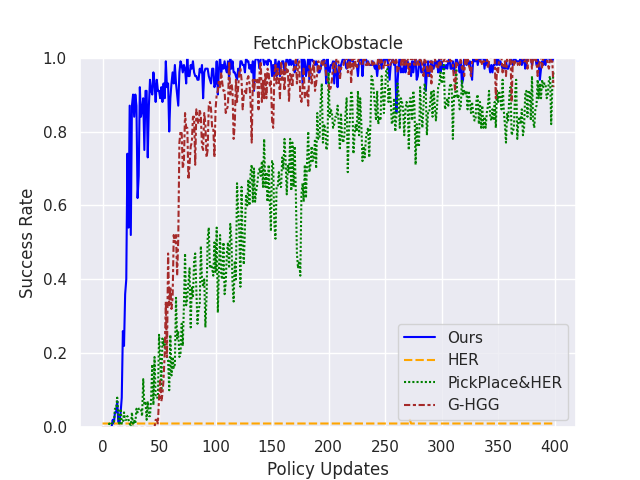}}
    \hfill
    \centering
    \subfloat[\label{3d}]{%
        \includegraphics[width=0.25\textwidth]{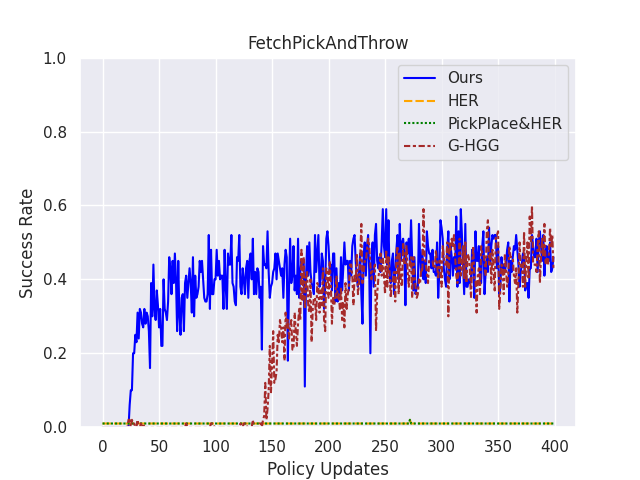}}
    \hfill
        \caption{Success rate  of the target policy $\pi(s, g)$ per each policy update on the different experiments described in Section \ref{sec:exp}.} 
        \label{fig:curves}
\end{figure*}

The results demonstrate that our method was able to accelerate the learning of HER, as it is capable of reaching a convergence point faster than HER (Fig~\ref{fig:curves}). 

In the case of the \textit{FetchPush} task, our method is able to obtain a success rate of 80\% in less than 20 policy updates, while it takes about 60 policy updates for HER to achieve the same performance. Our method is also able to achieve a 90\% success rate faster, as it only needs 25 policy updates, while HER requires approximately 70 updates (see Fig~\ref{3a}). A similar result is achieved for the \textit{FetchPickAndPlace} task, where the success rate of our method reaches 90\% accuracy using approximately 150 policy updates while HER takes about 180 policy updates, proving that our method increases the sample efficiency (see Fig~\ref{3b}). 

The results obtained for the different experiments for the \textit{FetchPickObstacle} task can be seen in Fig~\ref{3c}. While the task is too difficult for HER, which completely fails, our approach was able to reach an average of 95\% success rate in about 60 policy updates, compared to the approximate 175 updates needed for G-HGG. Furthermore, the average time per policy update of our model is 86 seconds, while for G-HGG it takes 117 seconds. The \textit{PickPlace\&HER} behaviour policy, while still working better than HER, proved to be inferior to the other methods explored, as the learning pace is slower. 

For the \textit{FetchPickAndThrow} experiments, we found that our method and G-HGG are able to achieve a 50\% success rate, while the other methods are unable to learn any useful policy (see Fig~\ref{3d}). After analysing the behaviour of the learned policy, it can be seen that for both methods the robot is observed to be capable throwing the block to the buckets of the first row nearest the robot base, but always fails when aiming for the second row buckets that are further away. This happens because during exploration the block almost never ends up in one of the second row buckets, making it impossible to generate hindsight goals for HER in these cases. Despite the similar results in terms of performance, our method is more sample efficient, as it converges faster than G-HGG, and more significantly, for each policy update our method takes 72.8 seconds on average, while the average time per update when using G-HGG is 396.5 seconds.

\section{Conclusion}

We have presented a novel algorithm, which we called Q-switch Mixture of Primitives Hindsight Experience Replay (QMP-HER), that improves the sample efficiency and the computation time of agents trained with HER by making use of primitive skills (i.e., policies) learned in simpler tasks. Instead of enforcing a curriculum for the behaviour policy using these primitives, the critic network of the target policy decides at each timestep during the exploration phase if the actions suggested by the primitive policies might be useful or not to solve the target task. 

Future work arising from this publication could include finding better primitives that replicate more specific behaviours, possibly by clustering environment transitions of simple tasks and replicating the behaviour of these clusters using any state-of-the-art behavioural modelling technique, such as decision transformers~\cite{dt} or trajectory transformers~\cite{tt}.

One of the weaknesses of the algorithm presented in this paper is the need for manually defined objectives and, therefore, another future direction could explore employing our method in conjunction with HGG, G-HGG as well as other algorithms that propose intermediate goals during the exploration phase, or even introducing a low-level parameter policy that determines how to instantiate  the primitives given an environment state as in~\cite{maple}.

\vspace{12pt}

\end{document}